\documentclass[sigconf]{acmart}

\AtBeginDocument{%
  }

\setcopyright{acmlicensed}
\copyrightyear{2018}
\acmYear{2018}
\acmDOI{XXXXXXX.XXXXXXX}

\acmConference[Conference acronym 'XX]{Make sure to enter the correct
  conference title from your rights confirmation email}{June 03--05,
  2018}{Woodstock, NY}

\acmISBN{978-1-4503-XXXX-X/2018/06}

\usepackage{float}
\usepackage{booktabs}
\usepackage{multirow} 
\usepackage{subfigure}
\usepackage{diagbox}
  
\usepackage{amsmath,amsthm,amssymb,amsfonts}
\usepackage{algorithm}
\usepackage{algorithmic}
\usepackage{pifont}
\usepackage{bbding}
\usepackage[most]{tcolorbox}

\begin{document}

\title{RealFactBench: A Benchmark for Evaluating Large Language Models in Real-World Fact-Checking}

\author{Shuo Yang}
\affiliation{
  \institution{The University of Hong Kong}
  \city{Hong Kong SAR}
  \country{China}}

\author{Yuqin Dai}
\affiliation{
  \institution{Tsinghua University}
  \city{Beijing}
  \country{China}}

\author{Guoqing Wang}
\affiliation{
  \institution{Ant Group}
  \city{Hangzhou}
  \country{China}}

\author{Xinran Zheng}
\affiliation{
 \institution{University College London}
 \city{London}
 \country{United Kingdom}}

\author{Jinfeng Xu}
\affiliation{%
  \institution{The University of Hong Kong}
  \city{Hong Kong SAR}
  \country{China}}

\author{Jinze Li}
\affiliation{
  \institution{The University of Hong Kong}
  \city{Hong Kong SAR}
  \country{China}}

\author{Zhenzhe Ying}
\affiliation{
  \institution{Ant Group}
  \city{Hangzhou}
  \country{China}}

\author{Weiqiang Wang}
\affiliation{
  \institution{Ant Group}
  \city{Hangzhou}
  \country{China}}

\author{Edith C.H. Ngai}
\affiliation{
  \institution{The University of Hong Kong}
  \city{Hong Kong SAR}
  \country{China}}

\renewcommand{\shortauthors}{Yang et al.}

\begin{abstract}
Large Language Models (LLMs) hold significant potential for advancing fact-checking by leveraging their capabilities in reasoning, evidence retrieval, and explanation generation. However, existing benchmarks fail to comprehensively evaluate LLMs and Multimodal Large Language Models (MLLMs) in realistic misinformation scenarios. To bridge this gap, we introduce RealFactBench, a comprehensive benchmark designed to assess the fact-checking capabilities of LLMs and MLLMs across diverse real-world tasks, including Knowledge Validation, Rumor Detection, and Event Verification. RealFactBench consists of 6K high-quality claims drawn from authoritative sources, encompassing multimodal content and diverse domains. Our evaluation framework further introduces the Unknown Rate (UnR) metric, enabling a more nuanced assessment of models' ability to handle uncertainty and balance between over-conservatism and over-confidence. Extensive experiments on 7 representative LLMs and 4 MLLMs reveal their limitations in real-world fact-checking and offer valuable insights for further research. RealFactBench is publicly available at Link\footnote{https://github.com/kalendsyang/RealFactBench.git}.

\end{abstract}



\maketitle

\section{Introduction}

In the era of information overload, false or misleading content is widely disseminated across multiple domains, including society, the economy, politics, and healthcare~\cite{augenstein2024factuality}. Owing to extensive reach, such misinformation can result in severe outcomes, such as inappropriate medication use~\cite{okati2024truth}, economic disruption~\cite{meel2020fake}, and reputational degradation~\cite{huan2024social}, underscoring the urgent need for reliable and scalable fact-checking systems. Traditional manual fact-checking via web searches cannot scale to the volume of online content, limiting its effectiveness for timely and comprehensive misinformation detection. Recent advances in LLMs have enabled automated fact-checking by combining reasoning, evidence grounding, and explanation generation~\cite{llmfcsurvey, pan2023fact}, showing strong potential to advance the field. Despite their potential, LLMs face significant challenges in handling real-world misinformation~\cite{rwchallenge, trendfact}. Unlike knowledge-based question-answering tasks, real-world rumors and events often involve multimodal content, evolving knowledge, and time-sensitive claims. These challenges highlight the importance of developing robust benchmarks to evaluate the fact-checking capabilities of LLMs and MLLMs in real-world scenarios. 

Several efforts have attempted to establish fact-checking benchmarks~\cite{factool, factscore, felm, pinocchio, factbench, longfact}. However, existing benchmarks suffer from the following limitations:
1) \textbf{Data Misrepresentation}: 
In real-world scenarios, data from news and social media is often multimodal, where images play a crucial role by providing fine-grained details such as specific locations, people, and events. However, most existing benchmarks~\cite{fever,factcheckbench,felm} remain text-only, limiting their ability to reflect the complexity of real-world inputs and to reliably evaluate model performance in practical settings.
2 ) \textbf{Evaluation Misrepresentation:} Current protocols struggle to evaluate models' handling of uncertainty. They fail to account for over-conservative or over-confident behaviors, leading to inflated or deflated performance metrics due to random guesses. 
3 ) \textbf{Scenario Misrepresentation}: Evaluations are often confined to static models, neglecting the critical role of external retrieval tools in fact-checking.
Therefore, integrating LLMs with external search tools is crucial~\cite{search}, as it enables access to up-to-date information and supports more realistic evaluation in dynamic real-world settings.

To address these limitations, we introduce RealFactBench, a comprehensive benchmark designed to evaluate the fact-checking capabilities of LLMs and MLLMs in real-world scenarios. Sourced from authoritative platforms and rigorously processed, RealFactBench comprises 6K data samples covering multimodal content and diverse domains such as politics, health, and science. It includes three tasks, ranging from static knowledge retrieval to dynamic misinformation analysis. To minimize random guessing, we allow models to respond with "Unknown" and propose the Unknown Rate (UnR) metric to assess over-conservatism or over-confidence behaviors. Evaluations across 7 LLMs and 4 MLLMs, including those equipped with web search tools, reveal that multimodal integration and real-time retrieval significantly enhance fact-checking performance. However, persistent challenges such as knowledge error and flawed reasoning highlight the need for further improvements in designing fact-checking systems for real-world applications.

Our key contributions can be summarized as follows: 
\begin{itemize}
    \item We propose RealFactBench, a benchmark to comprehensively evaluate LLMs and MLLMs on real-world fact-checking tasks, including Knowledge Validation, Rumor Detection, and Event Verification.

    \item We design a novel evaluation framework that assesses models across four dimensions: factual accuracy, prediction reliability, uncertainty handling, and explanation quality, with an additional focus on their performance in integrating web-based retrieval tools. 

    \item We release a dataset that incorporates real-world events, multimodal claims, and temporal dimensions, offering a robust foundation for testing models in practical misinformation scenarios.  

    \item Extensive evaluations of leading LLMs and MLLMs demonstrate the effectiveness of our benchmark, revealing actionable insights for improving fact-checking systems.
\end{itemize}
\section{Related Work}

\subsection{LLM-based Fact-Checking}

Fact-checking aims to assess the factuality of a claim, determining whether it is accurate or manipulated. Early pipeline-based methods~\cite{guo2022survey, Wadden2020, Saakyan2021} are limited by shallow retrieval corpora and poorly aligned with the verification task, resulting in unreliable outputs and reducing transparency. The advent of LLMs has revolutionized fact-checking systems~\cite{llmfcsurvey, pan2023fact, llmfc4, llmfc5}. By retaining and utilizing factual knowledge acquired during pretraining, LLMs can act as implicit knowledge bases~\cite{heinzerling2020language}. Recent works~\cite{llmfc2, llmfc3} have highlighted the potential of LLMs in combining implicit knowledge retrieval with reasoning, addressing many challenges encountered by traditional pipeline-based methods.

\subsection{Fact-Checking Benchmark}

The growing adoption of LLMs has spurred the development of benchmarks designed to evaluate their performance in fact-checking tasks. Early benchmarks primarily focused on text-based and closed-world question-answering (QA) tasks~\cite{pinocchio, halueval, selfaware, truthfulqa, factool}. For example, HaluEval\cite{halueval}, SelfAware\cite{selfaware}, and TruthfulQA~\cite{truthfulqa} emphasized short-text knowledge QA, targeting factual inaccuracies in concise claims. FactScore\cite{factscore} was designed for relatively simple biographical QA tasks, while FEVER~\cite{fever} extracted claims from Wikipedia for verification. To enhance factuality assessment, some benchmarks have introduced stricter protocols. FactCheck-Bench~\cite{factcheckbench} collected hallucinated responses from ChatGPT and provided multi-level document annotations for fine-grained evaluation. FACT-AUDIT~\cite{factaudit} proposed a dynamic multi-agent framework to evaluate rulings and arguments. FELM~\cite{felm} highlighted temporal generalization, emphasizing the challenges of adapting LLMs to evolving knowledge domains. Despite these advancements, most benchmarks remain constrained by their reliance on static or synthetic data, often focusing on single-modal claims. Recent studies~\cite{lrqfact,maft,mfcbench} have begun exploring multimodal fact-checking. MFC-Bench~\cite{mfcbench} emphasized the challenges of verifying visual misinformation, while MMFACKBench~\cite{mmfakebench} focused on detecting multimodal misinformation. However, these benchmarks still fall short in evaluating LLMs under real-world conditions, particularly in scenarios involving rumor propagation and time-sensitive events~\cite{factbench, trendfact}. This highlights the need for benchmarks based on real-world events that incorporate multimodal evidence to better assess factuality in practical scenarios. Additionally, fact-checking is inherently knowledge-intensive, requiring dynamic updates to reflect evolving information~\cite{intensive}. Existing evaluation protocols predominantly assess static LLMs, despite LLMs often lacking access to up-to-date or domain-specific information. Consequently, there is an objective need to integrate LLMs with external retrieval tools during evaluation. Such integration can enhance their ability to access timely and accurate information, thereby explicitly testing LLMs' ability in real-world environments. The detailed comparison of RealFactBench and existing fact-checking benchmarks can be found in Appendix~\ref{app: com}.
\section{RealFactBench}

\begin{figure*}
    \centering
    \includegraphics[width=1\linewidth]{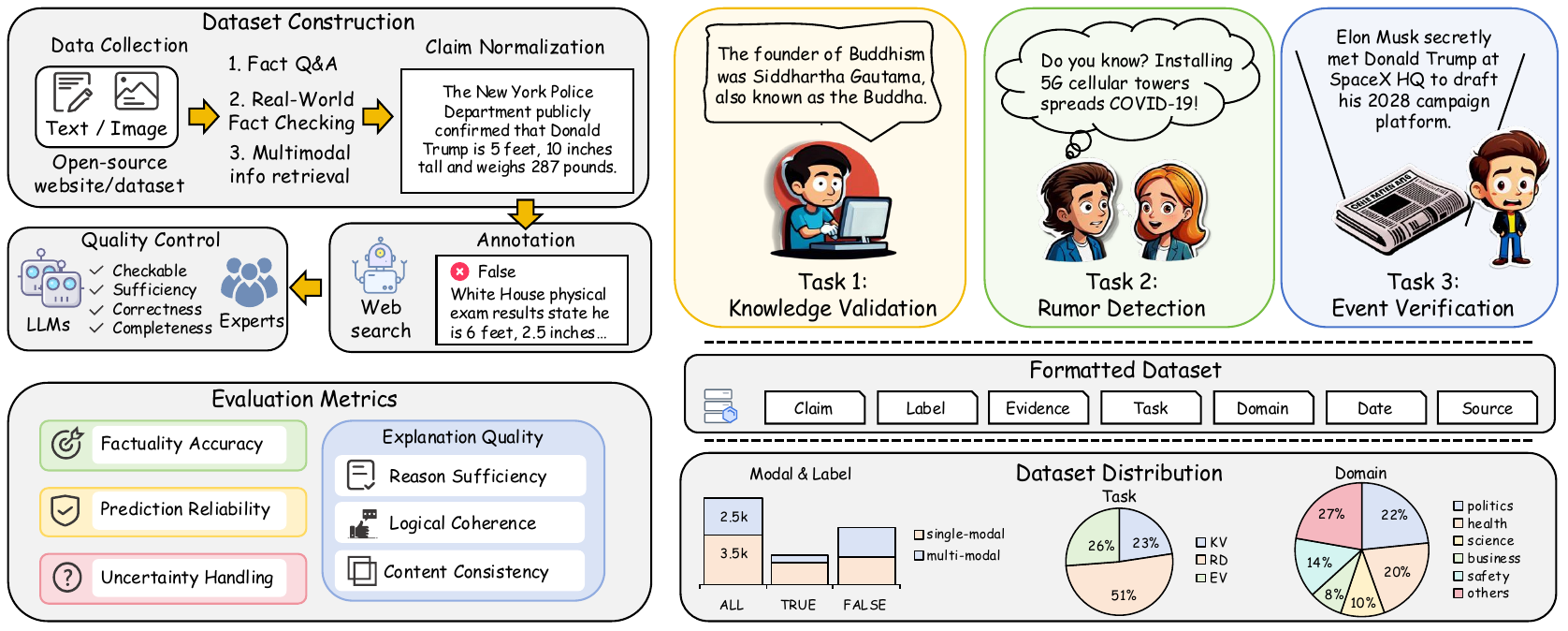}
    \vskip -0.15in
    \caption{Overview of RealFactBench. The left side illustrates the process of dataset construction and evaluation metrics. The right side displays the evaluation task, dataset format, and the detailed distribution of the dataset.}
    \label{fig: overview}
    \vskip -0.15in
\end{figure*}

In this section, we present RealFactBench, a comprehensive benchmark for evaluating the fact-checking capabilities of LLMs and MLLMs in real-world scenarios. An overview of RealFactBench is provided in Figure~\ref{fig: overview}. In contrast to existing benchmarks that rely on synthetic or closed datasets, RealFactBench incorporates verified data from authoritative and diverse sources, ensuring relevance, reliability, and practical applicability.

\subsection{Overview of Tasks}
RealFactBench comprises three specialized evaluation tasks: Knowledge Validation, Rumor Detection, and Event Verification. These tasks are illustrated in Figure~\ref{fig: overview}. Each task targets a specific fact-checking capability of models, from retrieving static knowledge to analyzing dynamic and complex misinformation.

\textbf{Knowledge Validation} focuses on evaluating the model's mastery of widely recognized and uncontested facts, such as scientific principles, historical data, and geographical information. This task tests the model’s ability to accurately retrieve and apply foundational knowledge, serving as a baseline for assessing its overall proficiency in factual accuracy.

\textbf{Rumor Detection} targets the identification and analysis of false or misleading information that spreads through public channels. This task requires the model to employ logical reasoning and critical thinking to detect inconsistencies, debunk manipulative narratives, and assess the credibility of unverified claims, thereby addressing challenges posed by misinformation in society.

\textbf{Event Verification} centers on fact-checking the accuracy of reported events, including news articles, social media posts, and official statements. Models must demonstrate the ability to gather information from multiple sources, evaluate source reliability, and conduct cross-verification to determine the truthfulness of the event in question. This task highlights the importance of handling dynamic, time-sensitive information in real-world contexts.

\subsection{Dataset Construction}

The construction of RealFactBench follows a systematic and rigorous pipeline to ensure the dataset's reliability, diversity, and applicability to real-world fact-checking scenarios. As illustrated in Figure~\ref{fig: overview}, the process consists of four key stages: data collection, claim normalization, annotation, and quality control.

\subsubsection{Data Collection}
RealFactBench integrates human-verified datasets with collected data from authoritative sources, ensuring coverage of static knowledge and dynamic fact-checking scenarios.

\textbf{Human-verified datasets.} We sampled claims from established public fact-checking datasets, including FELM~\cite{felm}, FactoolQA~\cite{factool}, Selfaware~\cite{selfaware}, FactCheck-Bench~\cite{factcheckbench}. These datasets provide a diverse collection of general knowledge claims verified by human annotators, serving as a foundation for evaluating models' ability to handle static factual information.

\textbf{Fact-checking websites.} We extracted claims from reputable fact-checking platforms such as Snopes, Science Feedback, and PolitiFact, which provide thorough analyses and evidence-backed judgments. A total of 1W+ claim samples spanning 2020-2025 were collected, with strict quality control yielding 4K+ high-quality samples. The examples primarily address rumors and event-related claims tied to political, health, and contemporary topics, ensuring alignment with dynamic real-world fact-checking challenges.

\textbf{Multimodality Source.} RealFactBench expands traditional fact-checking datasets by including multimodal data (e.g., textual claims paired with images, charts, or social media posts). Multimodal data was sourced from the Cosmos dataset~\cite{cosmos} and various archival websites such as Perma\footnote{https://perma.cc/}, ensuring a rich variety of multimedia resources.

\subsubsection{Data Processing}
The collected data undergoes a structured and rigorous processing pipeline to standardize its format, enhance usability, and ensure task relevance. The key steps are as follows:

\textbf{Normalization.} Claims are standardized into model-consumable statements to ensure consistency and enable uniform evaluation. Q\&A pairs (e.g., "What is the most common eye color in humans? Brown") are reformulated as declarative sentences ("The most common eye color in humans is brown"). Claims from websites or social media are retained in their original concise form.

\textbf{Annotation.} Each claim is annotated with a binary label indicating its factuality (TRUE/FALSE). To assess the explanatory ability of models, supporting evidence is retrieved from credible sources such as scientific studies, official reports, or reputable media platforms. We leverage LLMs to summarize the retrieved content, and the prompt template can be found in the Appendix~\ref{app: promptevidence}. Note that our focus is on the data that we collected. We have not undertaken redundant efforts to annotate claims sourced from public datasets, as these datasets have already undergone manual verification before their release.

For multimodal claims, alignment between textual and visual components is evaluated. If an image caption misrepresents the content of the image, the claim is labeled as False, with evidence describing the discrepancy. This rigorous annotation process ensures comprehensive evaluation of models' fact-checking capabilities, particularly in multimodal scenarios.

\textbf{Categorization:} Claims are categorized into one of three primary tasks: Knowledge Validation, Rumor Detection, or Event Verification. Categorization is based on the context and source of the claim. While some claims exhibit characteristics spanning multiple tasks, they are assigned to their primary category to streamline task-specific evaluations. This categorization enables targeted analysis of model performance across diverse fact-checking challenges.

\subsubsection{Quality Control}

Quality control is central to ensuring the accuracy and reliability of RealFactBench. The dataset benefits from multi-level verification processes, combining automated verification and expert review.

\textbf{Automated Verification.} Multiple LLMs\footnote{Here, we use DeepSeek-V3 and GPT-4-Turbo.} were used to evaluate the evidential grounding and fact-checking relevance of each claim. The models assessed sufficiency and integrity and provided justifications for claims deemed uncheckable. Claims consistently rated negatively were discarded. Additionally, independent annotators reviewed the claims, resolving disagreements via consensus. Detailed prompts and evaluation scenarios are presented in Appendix~\ref{app: promptquality}.

\textbf{Expert Verification.} A team of two doctoral students and one NLP specialist conducted a rigorous manual review of 1,000 randomly selected claims, assessing completeness and label accuracy while resolving anomalies to enhance dataset quality.

\subsection{Evaluation Metrics}

We propose a four-dimensional evaluation matrix consisting of Factual Accuracy, Prediction Reliability, Uncertainty Handling, and Explanation Quality to comprehensively assess model performance on RealFactBench, enabling a nuanced evaluation of model capabilities across diverse real-world fact-checking scenarios.

\textbf{Factual Accuracy: F1-Score (F1).}
Factual accuracy is critical for assessing a model's ability to classify claims as either true or false. We employ Macro F1-Score to offer a robust measure of a model's ability to accurately classify claims.

\textbf{Prediction Reliability: Matthews Correlation Coefficient (MCC).}
MCC provides deeper insights into the model's ability to consistently discern true and false claims. The output of MCC is a correlation coefficient between -1 and +1, where +1 indicates perfect classification, 0 represents performance equivalent to random guessing, and -1 denotes completely incorrect predictions.

\textbf{Uncertainty Handling: Unknown Rate (UnR).}
Real-world fact-checking often involves scenarios where models encounter uncertainty due to insufficient evidence or conflicting information. During evaluations, we observed that many LLMs output "Unknown" answers in such cases. To quantify this phenomenon, we introduce the Unknown Rate (UnR), defined as the proportion of claims for which the model outputs "Unknown." While a high UnR may indicate caution or a lack of knowledge about certain phenomena, it can also signal over-conservatism in predictions. Models designed for fact-checking must strike a balance between avoiding overly cautious behavior and maintaining accuracy. UnR serves as a valuable metric for analyzing these trade-offs and identifying tendencies toward uncertainty.

\textbf{Explanation Quality (EQ): Qualitative Analysis via LLM-as-Judge.} Providing robust explanations for fact-checking decisions is critical for improving the trustworthiness of model outputs. To evaluate Explanation Quality (EQ), we adopt the LLM-as-Judge framework~\cite{llmjudge}, which assesses model-generated explanations based on three dimensions: Content Consistency, Logical Coherence, and Evidence Sufficiency. Explanations are rated on a 0-10 scale, where a score of 0 is assigned to irrelevant explanations linked to incorrect predictions. The corresponding prompts and evaluation scenarios are detailed in Appendix~\ref{app: promptjudge}. Higher EQ scores indicate superior reasoning and enhanced interpretability, reflecting the model’s capacity to provide human-comprehensible justifications.

\subsection{Dataset Distribution}
As shown in Figure~\ref{fig: overview}, the dataset comprises 6K samples, with approximately 42\% containing multimodal input to better simulate complex real-world environments. The label distribution is imbalanced between TRUE and FALSE, mirroring the natural characteristics of fact-checking scenarios. The dataset supports three primary tasks: Knowledge Verification (KV, 23\%), Rumor Detection (RD, 51\%), and Event Verification (EV, 26\%), covering key applications from static knowledge validation to dynamic event verification and rumor identification. These tasks enable comprehensive evaluation of model performance across diverse contexts. Furthermore, the dataset spans multiple domains, including politics, health, science, business, and safety, ensuring broad applicability and addressing contemporary information challenges. By integrating multimodal data and diverse domains, the dataset provides a robust foundation for developing and evaluating advanced fact-checking systems.
\section{Experiment}

This section presents a thorough evaluation of various LLMs and MLLMs on RealFactBench. The experiments aim to benchmark the models' fact-checking capabilities across multiple tasks and scenarios, including the impact of web search tools, multimodal inputs, and knowledge cutoff. We also analyze failure cases to provide deeper insights into the limitations of current models.

\subsection{Experimental Settings}

We evaluated a diverse selection of state-of-the-art models on the benchmark tasks provided by RealFactBench. The models included in this study are Llama3.1-70B-Instruct~\cite{llama31}\footnote{To simplify, we will refer to Llama3.1-70B-I in the following text.}, Moonshot-V1~\cite{moonshot}, GPT-4o~\cite{gpt4}, GPT-4v~\cite{gpt4}, Gemini-2.5-Flash~\cite{genimi25}, Qwen-Plus~\cite{qwen}, and DeepSeek-V3~\cite{deepseek}, Claude-3.7-Sonnet~\cite{claude37}. To ensure fair and reproducible comparisons, all models were configured with a temperature of 0 and $top_n = 1$, which encourages deterministic outputs. All evaluations were conducted under a zero-shot prompting framework, where models were provided with task-specific instruction templates. These templates instructed the models to "act as a fact-checking expert" and explicitly required them to generate detailed reasoning for their judgments. The complete prompt template used in our experiments is provided in Appendix~\ref{app: prompteval}.

\begin{table*}[!ht]
\caption{Overall performance of LLMs on RealFactBench. $\uparrow$ / $\downarrow$ indicates that higher/lower is the better, respectively.}
\vskip -0.1in
\label{tab: overall}
\begin{tabular}{l|ccc|ccc|ccc|cccc}
\toprule
\multicolumn{1}{c|}{\multirow{2}{*}{Models}} & \multicolumn{3}{c|}{Knowledge   Validation} & \multicolumn{3}{c|}{Rumor Detection} & \multicolumn{3}{c|}{Event Verification} & \multicolumn{4}{c}{Overall}   \\ \cline{2-14} 
\multicolumn{1}{c|}{}   & F1{\footnotesize \%}$\uparrow$& MCC$\uparrow$& UnR{\footnotesize \%}$\downarrow$& F1{\footnotesize \%}$\uparrow$& MCC$\uparrow$& UnR{\footnotesize \%}$\downarrow$& F1{\footnotesize \%}$\uparrow$& MCC$\uparrow$& UnR{\footnotesize \%}$\downarrow$ & F1{\footnotesize \%}$\uparrow$& MCC$\uparrow$& UnR{\footnotesize \%}$\downarrow$ & EQ$\uparrow$ \\ \midrule
Llama-3.1-70B-I& 56.30  & 0.130& 4.14   & 55.39  & 0.210 & 18.31& 42.53& -0.082  & 35.75& 61.51  & 0.255 & 17.58  & 5.88 \\
Moonshot-V1   & 54.27  & 0.101 & \underline{3.53}   & 55.73  & 0.178 & \underline{8.77} & 52.01 & 0.040 & 20.50 & 63.53  & 0.301 & \underline{9.87}& 6.57\\
Gemini-2.0-Flash & 57.62  & 0.160& 3.84   & \underline{59.89}  & \underline{0.223} & 13.50& \underline{55.87} & \underline{0.122} & \underline{19.74} & \underline{68.93}  & \underline{0.380} & 11.54   & 7.01\\
Qwen-Plus & 55.39  & 0.129& \textbf{3.30}   & 52.89  & 0.099 & 20.75& 48.28& -0.028  & 30.89& 62.83  & 0.258 & 16.93   & 6.19 \\
GPT-4o& \underline{58.22}  & \underline{0.190} & 4.07   & 56.25  & 0.167 & 19.91& 49.01& -0.001  & 32.20& 64.50  & 0.292 & 17.24   & \underline{7.25}\\
DeepSeek-V3  & 56.54  & 0.133& 5.15   & 52.91  & 0.166 & 21.66& 43.91& -0.032  & 33.30& 61.09  & 0.254 & 18.58   & 5.96 \\
Claude-3.7-Sonnet & \textbf{60.24}  & \textbf{0.239}& 3.61   & \textbf{69.97}  & \textbf{0.402} & \textbf{6.48} & \textbf{63.10}& \textbf{0.262}& \textbf{14.80}& \textbf{73.48}  & \textbf{0.473 }& \textbf{7.57}& \textbf{7.52}\\ \bottomrule
\end{tabular}
\vskip -0.1in
\end{table*}

\subsection{Results on RealFactBench}

In Table \ref{tab: overall}, we present the overall performance of the evaluated models on RealFactBench. We draw the following conclusions:

\textbf{Overall Results across All Tasks.} Across all tasks, Claude-3.7-Sonnet demonstrates the strongest overall performance, achieving the highest scores on all three metrics. Gemini-2.0-Flash ranks second, with F1 scores of 68.93\%. These models benefit from their recent release dates, which afford access to up-to-date real-world knowledge and a balanced handling of uncertainty. In contrast, Moonshot-V1 exhibits overconfidence in uncertain scenarios, leading to misclassifications, while models such as Llama-3.1-70B-I and Deepseek-V3 tend to be overly conservative, missing opportunities for correct predictions. Performance also varies across task types: models generally excel in knowledge validation, where claims pertain to static or general facts, but struggle with event verification tasks, which exhibit high UoR and low MCC due to the scarcity of dynamic event data in training corpora. Finally, Claude-3.7-Sonnet and GPT-4o produce the most informative explanations, while Llama-3.1-70B-I underperforms in this regard, with an EQ score of 5.88/10 due to vague or incomplete outputs.

\textbf{Results on Knowledge Validation Task.} The knowledge validation task involves assessing claims against general or static facts. Despite asynchronous release timelines, the models' F1 in this task are relatively close. Notably, DeepSeek-V3 has the highest Uncertainty Ratio (UnR) at 5.15\%, indicating frequent use of the "Unknown" response due to insufficient confidence or evidence. GPT-4o achieves the second-highest performance after Claude-3.7-Sonnet, showcasing its robust capabilities in knowledge validation.

\textbf{Results on Rumor Detection Task.} The rumor detection task focuses on verifying the factuality of real-world rumors, often requiring dynamic reasoning and the ability to analyze conflicting information. Claude-3.7-Sonnet dominates this task, achieving an MCC of 0.402, which reflects its reliability in navigating ambiguous contexts with precise reasoning and confidence. In contrast, Moonshot-V1 demonstrates bias in its approach, labeling emotionally charged statements as rumors without sufficient supporting evidence. While this strategy yields a relatively high F1, it comes at the cost of a lower EQ score, as the model's judgments are based on emotional bias rather than factual analysis.

\textbf{Results on Event Verification Task.} The event verification task poses significant challenges due to its reliance on dynamic, real-world events and evolving evidence. Compared to knowledge validation, this task sees a marked increase in UnR across all models, emphasizing the need for up-to-date training data or external tools like web search. Older models such as Llama-3.1-70B-I and Qwen-Plus struggle notably, with negative MCC (-0.082 and -0.028, respectively), indicating poor alignment between predictions and ground truth. Their high UnR values (35.75\% and 30.89\%, respectively) reflect indecisiveness when dealing with event-based scenarios. These findings underscore the importance of incorporating temporal dependencies and real-time information into model training or inference pipelines. 

\subsection{Results w.r.t. Web Search Tool}

\begin{table}[t]
\caption{Performance comparison of LLMs with and without web search tools on RealFactBench.}
\vskip -0.1in
\label{tab: web}
\begin{tabular}{l|cccc}
\toprule
\multicolumn{1}{c|}{Models} & F1{\footnotesize \%}$\uparrow$& MCC$\uparrow$& UnR{\footnotesize \%}$\downarrow$ & CoR$\uparrow$ \\ \midrule
Moonshot-V1  & 63.53  & 0.301 & 9.87  &  -  \\
Moonshot-V1 with web & 86.96  & 0.742 & 0.53 &  60.07  \\ \midrule
GPT-4o       & 64.50  & 0.292 & 17.24  & - \\
GPT-4o with web   & 83.83 & 0.680      &  3.09   & 52.56   \\ \bottomrule
\end{tabular}
\vskip -0.1in
\end{table}

To examine the benefits of external resource access, we evaluated selected LLMs equipped with web search tools. The goal was to assess their ability to dynamically retrieve evidence and improve performance in real-world fact-checking. To control costs, testing was restricted to samples initially misclassified by the models. As shown in Table~\ref{tab: web}, web search access led to substantial performance gains.
Specifically, Moonshot-V1 achieved a 23.43\% increase in F1, while GPT-4o showed a 19.33\% improvement. Additionally, we calculated the Correction Rate (CoR), which measures the proportion of previously misclassified samples that were corrected using web search. Moonshot-V1 successfully corrected 60\% of its errors, whereas GPT-4o achieved a correction rate of 52.56\%. The disparity in CoR may stem from differences in search sources and tool quality. Additionally, Moonshot's advanced long-text processing capabilities allow it to maintain contextual coherence when handling complex content~\cite{moonshot}, resulting in more accurate search-based outputs and minimizing noise. These results underscore the importance of effective web search integration for enhancing LLM performance in dynamic fact-checking environments.

\subsection{Results w.r.t. Multimodal Input}

\begin{table}[]
\small
\caption{Performance comparison of MLLMs with and without multimodal information on RealFactBench.}
\vskip -0.1in
\label{tab: modal}
\begin{tabular}{l|ccc|ccc}
\toprule
\multicolumn{1}{c|}{\multirow{2}{*}{Models}} & \multicolumn{3}{c|}{Single-Modality} & \multicolumn{3}{c}{Multimodality} \\ \cline{2-7} 
\multicolumn{1}{c|}{}                        & F1{\footnotesize \%}$\uparrow$& MCC$\uparrow$& UnR{\footnotesize \%}$\downarrow$     & F1{\footnotesize \%}$\uparrow$& MCC$\uparrow$& UnR{\footnotesize \%}$\downarrow$    \\ \midrule
Gemini-2.0-Flash                             & 50.80       & 0.126     & 40.09       & 56.65      & 0.372    & 16.35      \\
GPT-4o                                       & 50.85       & 0.117     & 36.45       & 55.56      & 0.181    & 29.19      \\
Claude-3.7-Sonnet                            & 52.07       & 0.087     & 29.90       & 64.28      & 0.351    & 18.17      \\
GPT-4v                                       & -           & -         & -           & 66.82      & 0.407    & 15.21      \\ \bottomrule
\end{tabular}
\vskip -0.1in
\end{table}

In this experiment, we specifically evaluate the performance of MLLMs, such as Gemini-2.0-Flash, Claude-3.7-Sonnet, and GPT-4v, on RealFactBench's multimodal claims. These tasks involve claims that integrate textual and visual elements, requiring models to jointly process and reason across modalities. Images are provided as supplementary information to examine the models' abilities in image-text matching and image inference verification. From the results in Table~\ref{tab: modal}, GPT-4v achieved the best performance across all metrics, with an F1 score of 66.82\% and the lowest uncertainty ratio. Notably, multimodal information significantly improved the performance of Claude-3.7-Sonnet, resulting in a 12\% increase in F1 score over its single-modality counterpart. Additionally, multimodal fusion reduced overall model uncertainty, as evidenced by the 24\% decrease in UnR for Gemini-2.0-Flash. These results highlight the importance of leveraging multimodal inputs for fact-checking tasks, as incorporating visual elements not only enhances accuracy but also bolsters confidence in predictions by reducing uncertainty. 

\subsection{Impact of Knowledge Cutoff}

\begin{table}[]
\small
\caption{Performance comparison of LLMs, evaluated on samples before and after their knowledge cutoff dates.}
\vskip -0.1in
\begin{tabular}{l|ccc|ccc}
\toprule
\multicolumn{1}{c|}{\multirow{2}{*}{Models}} & \multicolumn{3}{c|}{Before} & \multicolumn{3}{c}{After} \\ \cline{2-7} 
\multicolumn{1}{c|}{}                        & F1{\footnotesize \%}$\uparrow$& MCC$\uparrow$& UnR{\footnotesize \%}$\downarrow$  &F1{\footnotesize \%}$\uparrow$& MCC$\uparrow$& UnR{\footnotesize \%}$\downarrow$ \\ \midrule
Llama-3.1-70B-I                                & 67.92   & 0.360  & 8.30     & 45.40  & -0.066 & 29.75   \\
Moonshot-V1                                  & 68.62   & 0.377  & 6.87     & 51.62  & 0.183  & 13.50    \\
Gemini-2.0-Flash                             & 69.80   & 0.396  & 8.30     & 54.23  & 0.085  & 19.25   \\
Qwen-Plus                                    & 64.80   & 0.296  & 11.12    & 51.92  & 0.040  & 20.50   \\
GPT-4o                                       & 70.07   & 0.405  & 7.53     & 48.74  & -0.022 & 29.00   \\
DeepSeek-V3                                  & 66.39   & 0.332  & 10.81    & 48.54  & 0.003  & 29.25   \\
Claude-3.7-Sonnet                            & 72.91   & 0.471  & 5.59     & 64.36  & 0.291  & 9.00    \\ \bottomrule
\end{tabular}
\vskip -0.1in
\end{table}

LLMs differ significantly in their knowledge cutoff dates, which directly affect their ability to evaluate claims about recent events or evolving topics. In this subsection, we analyze the impact of knowledge cutoff on performance. To conduct this analysis, we divided the dataset into two subsets based on the models' knowledge cutoff dates: the before subset, consisting of data samples from before 2024, and the after subset, containing data samples from after 2025. Our findings reveal a sharp decline in performance when models trained on earlier data are applied to the after subset. For instance, GPT-4o's accuracy dropped from 70.07\% on the before subset to 48.74\% on the after subset. This significant decrease underscores the critical role of up-to-date knowledge in fact-checking tasks and highlights the importance of incorporating real-world data for training.

\subsection{Failure Case Study}

Several common failure patterns were observed across the evaluated models, highlighting deficiencies in knowledge accuracy, reasoning, complex claim handling, and multimodal integration. Details are provided in Appendix~\ref{app: case}.

\textbf{Knowledge Error}: Training data inaccuracies often cause model failures by embedding false or outdated knowledge. For instance, Moonshot-V1 incorrectly attributed a quote to Winston Churchill, contradicting historical records. This highlights the susceptibility of static LLMs to misinformation within their training corpus. In contrast, with the aid of web search tools, the model dynamically retrieved authoritative evidence and refuted the claim, underscoring the value of access to current external knowledge sources.

\textbf{Flawed Reasoning}: Failures in reasoning occur when models correctly retrieve relevant information but fail to utilize it logically, especially in tasks requiring multi-step reasoning or probabilistic calculations. For instance, Qwen-Plus miscalculated survival rates by oversimplifying probabilistic deductions (e.g., "1 - fatality ratio"), ignoring context-specific factors such as age, comorbidities, and local variations in pandemic outcomes. This highlights reasoning weaknesses in scenarios that demand nuanced considerations or complex mathematical reasoning.

\textbf{Lost in Complex Text}: In cases involving intricate claims, models struggled to detect subtle inaccuracies embedded within otherwise plausible information. For instance, Llama-3.1-70B-I classified claims about Diatomaceous earth as "True" due to its general acceptance as aiding mineral absorption, failing to focus on the specific lack of scientific evidence supporting detoxification benefits, even though it already recognizes this. Models like DeepSeek-V3, however, managed to disentangle misleading components, demonstrating stronger capabilities in parsing nuanced claims.

\textbf{Misleading Information}: Multimodal reasoning errors were frequent among models processing both textual and visual inputs. For example, Gemini-2.0-Flash supported the claim of a rare purple lobster based solely on visual evidence presented in the manipulated image, without considering its authenticity. In contrast, Claude-3.7-Sonnet correctly identified visual tampering by referencing visible artifacts from digital alteration, showcasing its superior ability to integrate and reason about multimodal information critically.
\section{Conclusion}

In this paper, we introduce RealFactBench, a comprehensive benchmark designed to evaluate the fact-checking capabilities of LLMs and MLLMs. RealFactBench spans various domains and incorporates both single-modal and multimodal claims, supporting three primary tasks: Knowledge Validation, Rumor Detection, and Event Verification. These tasks cover core application scenarios in fact-checking, ranging from the accuracy of static knowledge to the truthfulness of dynamic events and the identification of rumor propagation. We propose a systematic framework for evaluating these models using a uniform prompt template, evaluating their factual accuracy, prediction reliability, uncertainty handling, and explanation quality. Our analysis of mainstream LLMs and MLLMs highlights critical limitations in knowledge accuracy, reasoning capabilities, and multimodal understanding. By identifying common failure patterns, we provide valuable insights and actionable directions for improving LLMs and MLLMs in real-world fact-checking tasks. RealFactBench aims to facilitate research into constructing more robust and reliable fact-checking systems. RealFactBench and its associated resources are publicly available to support further research in this field.

\begin{acks}
This work was supported by Ant Group Research Intern Program.
\end{acks}

\bibliographystyle{ACM-Reference-Format}
\bibliography{reference}

\newpage
\appendix
\section{Benchmark Comparison}
\label{app: com}

\begin{table*}[htbp]
\caption{Comparison of benchmarks related to fact-checking.}
\vskip -0.1in
\label{tab: com}
\begin{tabular}{l|c|c|c|c|c|c}
\toprule
Benchmarks & \# Claim & Source & Multimodality & Real-world Fact & Uncertainty Evaluation & Web Search Evaluation \\ \midrule
FELM~\cite{felm} & 847& SD& \ding{55}& \ding{55} & \ding{55} & \ding{55}\\
Factcheck-Bench~\cite{factcheckbench} & 94 & SD& \ding{55}& \ding{55} & \ding{55} & \ding{55}\\
FactScore~\cite{factscore}& 500& SD& \ding{55}& \ding{55} & \ding{55} & \ding{55}\\
FactTool~\cite{factool}& 539& SD& \ding{55}& \ding{55} & \ding{55} & \ding{55}\\
Pinochio~\cite{pinocchio}& 20K& ED& \ding{55}&\ding{55}& \ding{55}& \ding{55}\\
LongFact~\cite{longfact}& 2K & SD& \ding{55}& \ding{55} & \ding{55} & \ding{55}\\
FactBench~\cite{factbench} & 1K & SD& \ding{55}& \ding{55} & \ding{55} & \ding{55}\\
MFC-Bench~\cite{mfcbench}& 35K& ED& \ding{52}& \ding{52} & \ding{55} & \ding{55}\\
MMFakeBench~\cite{mmfakebench}& 11K& SD& \ding{52}&\ding{52}& \ding{55} & \ding{55}\\
TrendFact~\cite{trendfact}& 7.6K& RC& \ding{55}& \ding{52} & \ding{55} & \ding{55}\\ \midrule
RealFactCheck & 6K & RC& \ding{52}& \ding{52} & \ding{52} & \ding{52}\\ \bottomrule

\multicolumn{7}{l}{SD: Synthetic Data, ED: Existing Dataset, RC: Real-world Collection}
\end{tabular}
\vskip -0.1in
\end{table*}

Table~\ref{tab: com} compares RealFactBench with existing fact-checking benchmarks based on the number of claims, data sources, and evaluation protocol. Most existing benchmarks primarily rely on synthetic data (SD) or existing datasets (ED), whereas RealFactBench is constructed from real-world collections (RC). Furthermore, RealFactBench is the only benchmark that simultaneously supports multimodality, real-world fact evaluation, uncertainty evaluation, and web search-based testing, offering significantly broader coverage compared to prior works. While other benchmarks vary in scale, with claim counts ranging from 94 to 35K, RealFactBench achieves a balance between data scale (6K claims) and rich evaluation functionality, making it a comprehensive and practical benchmark for fact-checking tasks. 

\section{Prompt Template}

\subsection{Prompt for Summarizing Evidence}
\label{app: promptevidence}

\begin{tcolorbox}[breakable, colback=gray!10, colframe=gray!60, sharp corners=southwest, rounded corners=northwest]
As a professional evidence generator, your task is to generate summarized evidence from the judgment content in response to the given claim and label.

\setlength{\parindent}{0.5em}- Evidence SHOULD directly support the labels of the claims and be relevant to the judgment content.

\setlength{\parindent}{0.5em}- DO NOT add any information that is not contained in the judgment content.

\setlength{\parindent}{0em}Your response MUST strictly follow this format:

\setlength{\parindent}{0.5em}\{

\setlength{\parindent}{2em}"evidence": "extract evidence from the judgment content that supports the  label (maximum 3 points)",

\setlength{\parindent}{0.5em}\}

\setlength{\parindent}{0em}Now process this input:
\end{tcolorbox}

\subsection{Prompt for Automated Quality Control}
\label{app: promptquality}

\begin{tcolorbox}[breakable, enhanced, colback=gray!10, colframe=gray!60, sharp corners=southwest, rounded corners=northwest]
As a professional claim quality inspector. Given a claim, its label, and supporting evidence, your task is to determine whether the claim has the value of fact-checking and whether the evidence is sufficient to support the label.

\setlength{\parindent}{0.5em}- A checkable claim can be verified or disproven with evidence or reliable sources, especially if it involves public interest, controversy, or questionable origins.

\setlength{\parindent}{0.5em}- Sufficient evidence can be used to support the label of the claim, and it should be relevant to the claim and provide clear reasoning for the label.

\setlength{\parindent}{0em}Your response MUST strictly follow this format:

\setlength{\parindent}{0.5em}\{
    
    \setlength{\parindent}{2em}"checkable": "true if the claim is checkable, false otherwise",
    
    \setlength{\parindent}{2em}"sufficient": "true if the evidence is sufficient to support the label, false otherwise",
    
    \setlength{\parindent}{2em}"explanation": "a clear and concise explanation of why the claim is checkable or not, and why the evidence is sufficient or not.",

\setlength{\parindent}{0.5em}\}

\setlength{\parindent}{0em}Now process this input:
\end{tcolorbox}

\subsection{Prompt for Explanation Quality Metrics}
\label{app: promptjudge}

\begin{tcolorbox}[colback=gray!10, colframe=gray!60, sharp corners=southwest, rounded corners=northwest]
As a professional fact-checking system evaluator, your task is to holistically evaluate an AI assistant’s response against gold standards. The evaluation should be based on three dimensions: Content Consistency, Logical Coherence, and Reason Sufficiency.

1. Content Consistency: Does the explanation consistently with the claim and evidence provided?

2. Logical Coherence: Is the explanation logically structured and free of contradictions?

3. Reason Sufficiency: Are the sources or reasoning provided adequate to justify its factuality judgment?

\setlength{\parindent}{0.5em} - If the verdict is incorrect, the score cannot exceed 4.

\setlength{\parindent}{0.5em} - If the verdict is correct but the reasoning does not fully align with the gold standard, the score range is 5-7.

\setlength{\parindent}{0.5em} - If the verdict is correct and reasoning fully aligns with the gold standard, the score range is 8-10.

\setlength{\parindent}{0em}Your response MUST strictly follow this format:

\setlength{\parindent}{0.5em}\{

\setlength{\parindent}{2em} "score": "evaluation score",
    
\setlength{\parindent}{2em} "justification": "justification for scoring"
    
\setlength{\parindent}{0.5em}\}

\setlength{\parindent}{0em}Now process this input:
\end{tcolorbox}

\subsection{Prompt for Model Evaluation}
\label{app: prompteval}
\begin{tcolorbox}[colback=gray!10, colframe=gray!60, sharp corners=southwest, rounded corners=northwest]
As a professional fact-checking assistant, your task is to check the factuality of a given claim.

\setlength{\parindent}{0.5em}- ALWAYS give a clear, decisive "true" or "false" verdict if evidence strongly supports it. 

\setlength{\parindent}{0.5em}- ONLY use "unknown" if no evidence is available to verify or refute the claim, and explain why in your reason.

\setlength{\parindent}{0em}Your response MUST strictly follow this format:

\setlength{\parindent}{0.5em}\{

    \setlength{\parindent}{2em}"verdict": "true" / "false" / "unknown",
    
    \setlength{\parindent}{2em}"reason": "a concise explanation (maximum 3 points) citing specific evidence"
    
\setlength{\parindent}{0.5em}\}

\setlength{\parindent}{0em}Now process this input:
\end{tcolorbox}

\section{Case Study}
\label{app: case}

\begin{figure*}[t]
    \centering
    \includegraphics[width=1\linewidth]{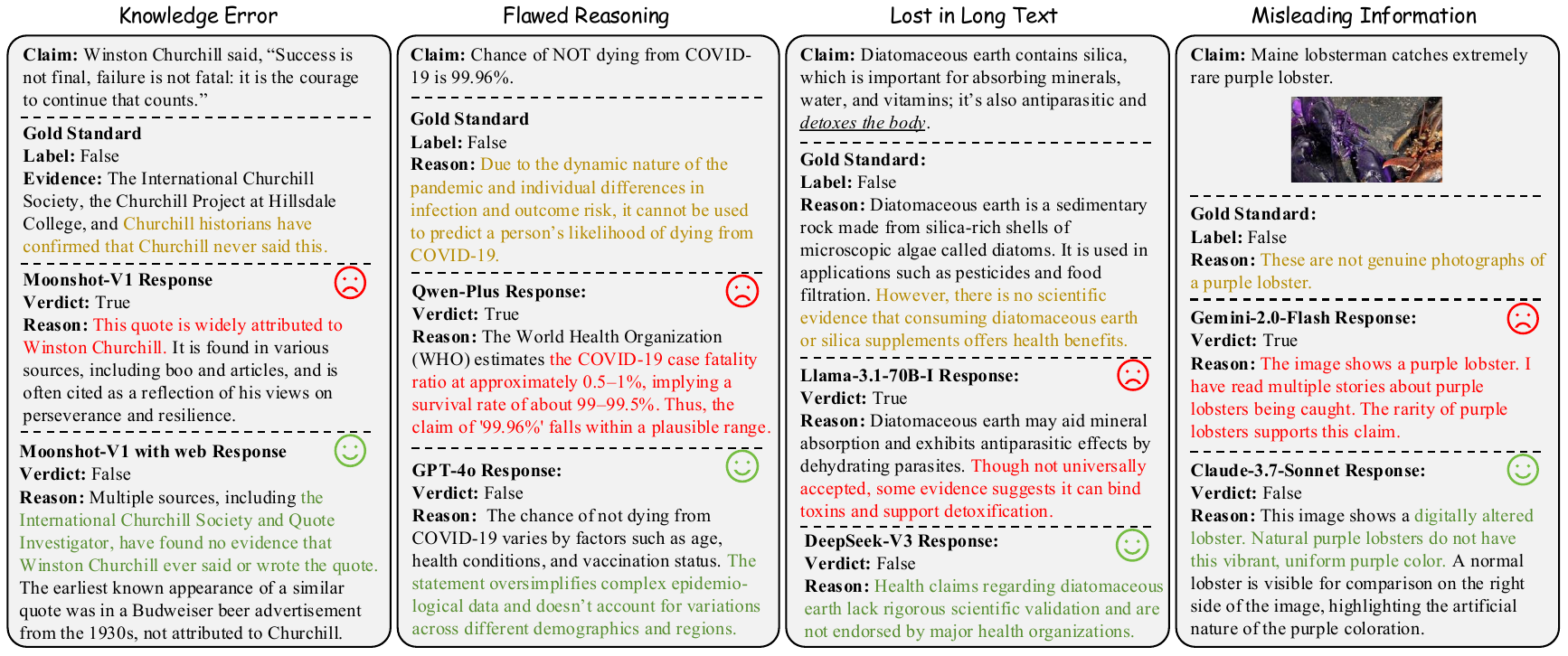}
    \caption{Case Study. Examples of claims along with their gold-standard answers and model responses. Green: Supporting evidence provided by RealFactBench. Red: Incorrect reasoning by the model. Brown: Correct reasoning by the model.}
    \label{fig: case}
\end{figure*}

Figure~\ref{fig: case} presents examples of four failure cases in fact-checking: Knowledge Error, Flawed Reasoning, Lost in Long Text, and Misleading Information. Each example includes the gold standard label, evidence, and model responses, with correct and incorrect outcomes marked by green and red symbols, respectively. The examples highlight distinct challenges encountered by fact-checking models: reliance on inaccurate yet widely attributed information (Knowledge Error), oversimplification of nuanced reasoning (Flawed Reasoning), difficulty in comprehending and e valuating detailed textual context (Lost in Long Text), and inability to detect manipulated or fabricated content (Misleading Information). These examples illustrate the diverse error patterns in current models and highlight the need for more robust fact-checking capabilities.

\section{Cost Analysis}

\begin{table}[!h]
\caption{Approximate number of tokens for Prompt and LLM response for each sample.}
\label{tab: token}
\begin{tabular}{l|c|c|c}
\toprule
\multicolumn{1}{c|}{Operations} & Input & Output & Overall \\ \midrule
Evidence Summary            & 1200  & 300    & 1500    \\
Quality Control            & 500   & 100    & 600     \\
Single-modal Test          & 150   & 150    & 300     \\
Multimodal Test           & 700   & 150    & 850     \\
Test with Web Search       & 10000 & 150    & 10150   \\
EQ Judgment               & 500   & 50     & 550     \\ \bottomrule
\end{tabular}
\end{table}

The primary costs of RealFactBench stem from the dataset construction phase, which involved significant expenses for manual data collection and expert annotation, with all contributors fairly compensated to ensure transparency and avoid any disputes. Additional costs were incurred through the deployment of open-source models like LLaMA and DeepSeek-V3, as well as API calls to commercial services for performance evaluation. As shown in the Table~\ref{tab: token}, the estimated token counts for each sample across tasks reveal that web search incurs the highest cost among all operations. 

\section{Ethical Considerations}

All data in RealFactBench is sourced from publicly accessible information, and we have adhered to a responsible data collection protocol. We explicitly commit to using the collected data solely for academic and research purposes, ensuring compliance with ethical research norms. Furthermore, any images or text included in the dataset remain the intellectual property of their original authors, and their inclusion in RealFactBench is intended only for advancing research into combating misinformation, without infringing on the rights of content creators. If there are any copyright-related concerns or requests, please feel free to contact us, and we will respond promptly and address them appropriately.

Despite careful sourcing, the dataset may reflect biases inherent in publicly available content, including political, cultural, or societal biases. These biases could propagate through fact-checking models trained or tested on RealFactBench, leading to unfair or inaccurate outputs. We encourage users of the benchmark to critically analyze and document such biases, promoting fairness and transparency in their research and applications.

\end{document}